\documentclass[11pt]{article}

\usepackage{acl}

\usepackage[T1]{fontenc}
\usepackage[utf8]{inputenc}
\usepackage{times}
\usepackage{latexsym}
\usepackage{microtype}
\usepackage{inconsolata}

\usepackage{amsmath,amssymb,amsfonts}
\usepackage{graphicx}
\usepackage{booktabs}
\usepackage{multirow}
\usepackage{array}
\usepackage{xcolor}
\usepackage{enumitem}
\setlist{nosep}
\usepackage{threeparttable}

\newcommand{\kg}{PrimeKG}

\title{MHGraphBench: Knowledge Graph-Grounded Benchmarking of Mental Health Knowledge in Large Language Models}

\author{
\begin{tabular}{c}
\textbf{Weixin Liu}$^{1}$ \quad
\textbf{Congning Ni}$^{2}$ \quad
\textbf{Shelagh A. Mulvaney}$^{1}$ \quad
\textbf{Susannah L. Rose}$^{2}$ \\
\textbf{Murat Kantarcioglu}$^{3}$ \quad
\textbf{Bradley A. Malin}$^{1,2}$ \quad
\textbf{Zhijun Yin}$^{1,2}$ \\
$^{1}$ Vanderbilt University, Nashville, TN, USA \\
$^{2}$ Vanderbilt University Medical Center, Nashville, TN, USA \\
$^{3}$ Virginia Tech, Blacksburg, VA, USA \\
\texttt{\{weixin.liu, shelagh.mulvaney\}@vanderbilt.edu} \\
\texttt{\{congning.ni.1, susannah.rose, b.malin, zhijun.yin\}@vumc.org} \\
\texttt{muratk@vt.edu}
\end{tabular}
}

\begin{document}
\maketitle

\begin{abstract}
Large language models (LLMs) are increasingly used in the mental health domain, yet it remains unclear how well they capture related biomedical knowledge and how reliably they apply it to clinically salient structured judgments. Here, we present a knowledge-graph (KG)-grounded benchmark for assessing LLMs on mental-health entity recognition, relation judgment, and two-hop reasoning. The benchmark is derived from \kg{} and comprises nine task families with KG-supported answers and controlled negative options. % Beyond accuracy, we quantify graph-wide coverage over entities, relations, and triples, and provide fine-grained entity- and relation-centric analyses. 
Experiments across 15 closed- and open-source LLMs reveal a persistent recognition-to-judgment gap: leading models achieve near-ceiling performance on entity typing and on the small relation-typing subset, yet they still struggle with relation prediction and two-hop reasoning. %, especially for clinically sensitive relation tasks such as \texttt{contraindication}. 
Additionally, short KG-derived snippets benefit some models but degrade performance for others. Moreover, output-format reliability can substantially influence measured performance under constrained multiple-choice settings, highlighting the critical role of response validity in benchmark-based evaluation. MHGraphBench should therefore be interpreted as evaluating agreement with a curated mental-health slice of \kg{} under a constrained multiple-choice interface, rather than as a direct assessment of real-world clinical safety.
\end{abstract}

% =========================================================
\section{Introduction}
\label{sec:intro}

Mental health disorders impose a large and growing burden worldwide~\cite{gbd2022global}. Clinical care and translational research in mental health often require integrating heterogeneous biomedical evidence, including disorder relationships, phenotypes and exposures, medication-use boundaries (e.g., \texttt{indication} vs.\ \texttt{contraindication} vs.\ \texttt{off-label use}), and disease-associated biological signals~\cite{freidel2025knowledge, gao2025large, kyrios2024off, rosland2025off}. These characteristics make mental-health applications particularly sensitive not only to whether models can recognize relevant biomedical entities, but also to whether they can correctly apply knowledge to clinically salient structured judgments.

Large language models (LLMs) have shown strong performance in biomedical and clinical tasks and have attracted growing interest in healthcare applications~\cite{singhal2025toward, saab2024capabilities, iqbal2025impact, li2024chatgpt}, including mental-health settings~\cite{volkmer2024large, obradovich2024opportunities}. Most existing evaluations still report aggregate accuracy on broad biomedical or clinical benchmarks, offering limited insight into two questions that are especially important in mental health: (i) how broadly an LLM covers mental-health biomedical knowledge, and (ii) whether it can reliably distinguish clinically salient and safety-sensitive relation boundaries~\cite{arora2025healthbench, cai2024medbench}. 

At the same time, mental-health-specific evaluation is evolving rapidly. Recent benchmarks have begun to assess psychiatric diagnostic decision-making, realistic counseling and help-seeking interactions, and trustworthiness in mental-health settings~\cite{song2026mentalbench, xiong2026trustmh, li2025counselbench}. These efforts broaden the scope of mental-health LLM evaluation, but they primarily emphasize diagnosis, counseling quality, or trustworthiness rather than verifiable structured biomedical knowledge and knowledge-graph (KG)-grounded relation reasoning. This challenge is further complicated by growing evidence that multiple-choice LLM evaluation can itself be sensitive to option ordering, prompt formatting, constrained answer formats, and output parsing rules~\cite{wang2024beyond, pezeshkpour2024large, zheng2023large}. Accordingly, our goal is not to evaluate real-world clinical decision-making or clinical safety directly, but rather to evaluate KG-grounded structured discrimination and short-path reasoning with respect to a curated mental-health graph under a constrained multiple-choice interface.

KGs provide curated biomedical facts in a structured and machine-verifiable form. They are well-suited to benchmark construction because they support automatic QA generation from factual triples, systematic negative sampling, and interpretable task design over entities, relations, and paths~\cite{chandak2023building, sun2023think, salnikov2023large, markowitz2025kg}. Biomedical KGs such as \kg{} also illustrate the value of graph-structured resources for downstream biomedical reasoning and analysis~\cite{chandak2023building}. For mental health, KG-grounded evaluation is especially useful because it enables controlled benchmarking over clinically salient relation families rather than relying only on open-ended prompting. In addition, benchmark items derived directly from KG facts are verifiable against the underlying graph, enabling analysis not only of task accuracy but also of graph-wide knowledge coverage.

\begin{figure*}[t]
\centering
\includegraphics[width=0.8\textwidth]{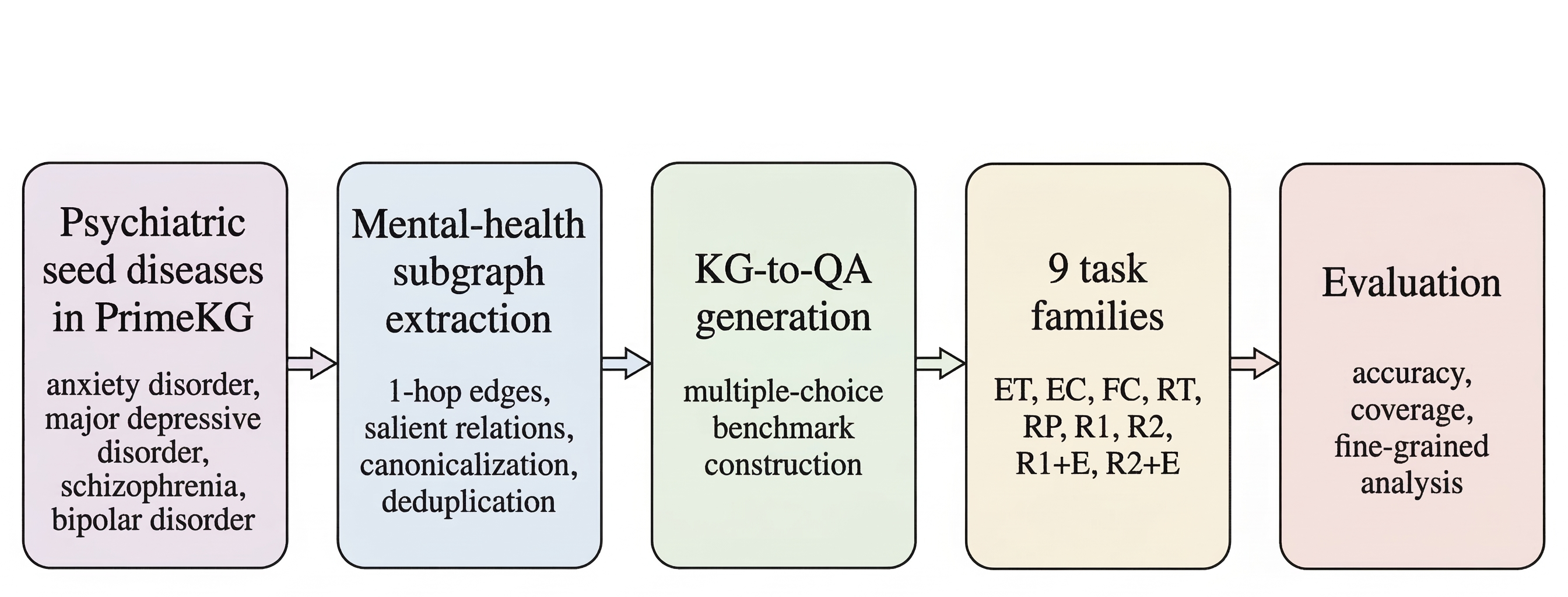}
\caption{Overview of the KG-grounded mental-health benchmark framework. Starting from 42 final psychiatric seed disease nodes in \kg{}, we extract a clinically focused mental-health subgraph, transform the resulting knowledge graph into a multiple-choice question-answering (QA) benchmark with nine task families, and evaluate models using accuracy, coverage, and fine-grained analyses.}
\label{fig:framework}
\end{figure*}

In this paper, we introduce \textbf{MHGraphBench}, a KG-grounded benchmark for evaluating mental-health biomedical knowledge in LLMs using a curated mental-health subgraph of \kg{}. We define the benchmark domain with 42 psychiatric seed disease nodes, extract a clinically focused subgraph, and transform it into nine standardized multiple-choice task families spanning entity recognition, relation judgment, and short disease-mediated reasoning. All benchmark items are derived from KG-backed facts with controlled negatives, making MHGraphBench a structured and reproducible benchmark for evaluating mental-health biomedical knowledge with respect to a curated graph rather than a direct measure of broader clinical reasoning or real-world clinical safety. Beyond benchmark accuracy, we also quantify graph-wide coverage over entities, relations, and triples, and provide fine-grained entity- and relation-centric analyses to localize where models succeed or fail. Figure~\ref{fig:framework} summarizes the overall pipeline, including psychiatric seed selection, mental-health subgraph extraction, KG-to-QA generation, task construction, and evaluation.

\paragraph{Design principle: verifiable KG-grounded evaluation.}
Our central design principle is that benchmark items should be automatically derived from KG facts, paired with explicit negative sampling, and remain verifiable against the underlying mental-health subgraph. This makes the evaluation scalable and reproducible while also allowing us to analyze which entities, relations, and graph regions models handle well or poorly, rather than summarizing performance only with a single overall accuracy number.

Using MHGraphBench, we ask four main questions: 1) Do models that perform well on entity typing and on the small relation-typing subset also perform well on clinically meaningful relation judgment? 2) How difficult is short disease-mediated reasoning relative to simpler recognition tasks? 3) What additional insight do graph-wide coverage and fine-grained analyses provide beyond average task accuracy? 4) When short KG-derived evidence is added, does it consistently help model performance?

Our experiments across 15 models yield three main takeaways. First, even the strongest models are near ceiling on entity typing and on the small relation-typing subset but remain substantially weaker on relation prediction and two-hop reasoning, revealing a persistent recognition-to-judgment gap. Second, clinically sensitive relation families, especially \texttt{contraindication}, remain difficult across models, and open-source models lag well behind the strongest GPT-series systems on the overall benchmark. Third, graph-wide coverage and evidence augmentation provide complementary insight: coverage rankings do not fully match average task rankings, and short KG-derived evidence helps some models but degrades others.

\paragraph{Contributions}
\begin{itemize}[leftmargin=*]
    \item We construct a mental-health benchmark from a curated \kg{} subgraph defined by 42 psychiatric seed disease nodes. It includes nine standardized multiple-choice task families spanning entity recognition, relation judgment, and short two-hop reasoning, all with KG-supported ground truth and controlled negatives.
    \item We introduce graph-wide coverage metrics and fine-grained entity- and relation-centric analyses to complement raw task accuracy and localize model strengths and weaknesses within the mental-health graph.
    \item We present empirical results across 15 LLMs showing a persistent recognition-to-judgment gap, mixed effects of evidence augmentation, and the importance of response-format reliability in constrained benchmark evaluation.
\end{itemize}

% =========================================================
\section{Related Work}
\label{sec:related}

Several studies evaluate LLMs on biomedical question answering, clinical reasoning, factuality, expert-style exam tasks, and broader healthcare use cases~\cite{singhal2025toward, saab2024capabilities, iqbal2025impact, li2024chatgpt}. These benchmarks provide useful broad capability signals, but they often report aggregate scores over heterogeneous tasks and therefore offer limited insight into mental-health-specific knowledge or failure modes~\cite{singhal2025toward, saab2024capabilities, arora2025healthbench}. In addition, prior work has shown that multiple-choice LLM evaluation can itself be sensitive to factors such as option ordering, prompt formatting, constrained answer formats, and output parsing rules~\cite{pezeshkpour2024large, zheng2023large, wang2024beyond}.

Knowledge graphs (KGs) have been used to probe factual knowledge, generate verifiable benchmarks, and study model reasoning behavior under controlled perturbations~\cite{chandak2023building, sun2023think, salnikov2023large, markowitz2025kg}. Biomedical KG resources such as the Drug Repurposing Knowledge Graph (DRKG) and \kg{} further illustrate the value of structured graph representations for integrating heterogeneous biomedical evidence~\cite{ioannidis2020drkg, chandak2023building}. KG-grounded benchmarks are especially appealing because they support scalable question generation, controlled negative sampling, explicit gold labels, and interpretable evaluation over entities, relations, and paths. However, relatively little prior work has focused on mental-health-centered KG benchmarking with clinically salient relation boundaries, short-path reasoning tasks, and graph-level coverage analysis.

Mental-health biomedical knowledge spans disorder relationships, medication-use boundaries, phenotypes, exposures, and biological associations~\cite{freidel2025knowledge, gao2025large}. Recent mental-health-specific benchmarks extend evaluation beyond broad biomedical or clinical QA by targeting psychiatric diagnostic decision-making, counseling and help-seeking quality, and trustworthiness in safety-sensitive settings~\cite{song2026mentalbench, xiong2026trustmh, li2025counselbench}. These benchmarks broaden the scope of mental-health LLM evaluation, but they primarily emphasize diagnosis, counseling quality, or trustworthiness rather than verifiable structured biomedical knowledge and KG-grounded relation reasoning. Our benchmark complements these efforts by focusing on a curated mental-health slice of \kg{} and evaluating entity recognition, relation judgment, short reasoning behavior, and graph-wide coverage in a unified KG-grounded framework.
% =========================================================
\section{Benchmark Construction and KG-to-QA Generation}
\label{sec:dataset_method}

Figure~\ref{fig:framework} summarizes the end-to-end pipeline of the proposed framework. Psychiatric seed diseases define the target domain, subgraph extraction yields a curated mental-health slice of \kg{}, KG-to-QA generation converts graph facts into benchmark items, and evaluation reports aggregate accuracy, graph-wide coverage, and fine-grained analyses.

\subsection{Mental-Health Subgraph}
\label{subsec:primekg}

\kg{} is a large, publicly available biomedical knowledge graph that integrates curated associations across drugs, diseases, genes/proteins, pathways, and other biomedical entities, in which typed nodes are connected by semantically defined relation edges~\cite{chandak2023building}. In PrimeKG, disease nodes are encoded using terms from the Mondo Disease Ontology (MONDO) and grouped into clinically meaningful disease nodes during graph construction~\cite{chandak2023building}. Building on this disease layer, we manually curated a high-precision candidate seed list of 44 PrimeKG disease nodes with psychiatric relevance. We then excluded two candidates during post-curation: \emph{X-linked intellectual disability-psychosis-macroorchidism syndrome}, which was considered outside the intended benchmark scope, and \emph{multiple personality disorder}, which was considered outdated terminology. This yielded 42 final psychiatric seed disease nodes (see Appendix~\ref{app:seed_list}). This final seed set defines the benchmark's mental-health domain boundary and provides a reproducible basis for subgraph extraction. Starting from these 42 final psychiatric seed disease nodes, we extracted all 1-hop seed-touching edges and then retained only a fixed set of clinically salient relation families, including drug--disease usage relations (\texttt{indication}, \texttt{contraindication}, \texttt{off-label use}), disease--disease links, and related biomedical associations such as \texttt{disease\_protein}, \texttt{disease\_phenotype\_positive}, and \texttt{exposure\_disease}.

This procedure yields 9{,}242 raw edges connecting to the seed disease nodes. We canonicalized each retained relation to a consistent head/tail type signature, with symmetric handling for \texttt{disease\_disease}, and deduplicated triples after canonicalization. The resulting mental-health subgraph contains 4{,}621 unique triples over 1{,}847 entities and 7 retained relation types, serving as the sole source of benchmark ground truth.

\subsection{KG-to-QA Task Suite}
\label{subsec:task_gen}

From the curated \kg{} mental-health subgraph, we generate nine standardized multiple-choice tasks with letter-only outputs and KG-grounded answers:

\begin{itemize}[leftmargin=*]
    \item \textbf{Entity Typing (ET)} asks the model to identify the type of a target entity, using the entity type in the subgraph as the gold label.
    \item \textbf{Entity Clustering (EC)} presents an ``odd-one-out'' problem formed by sampling four entities of the same type and one entity of a different type.
    \item \textbf{Fact Checking (FC)} asks whether a candidate triple is supported by the subgraph. Negative examples are generated by replacing the head or tail entity with a type-matched alternative under the same relation and retaining only perturbed triples that are unsupported by the extracted subgraph. FC instances are balanced per relation so that each relation contributes equal numbers of ``Yes'' and ``No'' examples.
    \item \textbf{Relation Typing (RT)} asks the model to identify the correct head$\to$tail type-pair schema of a relation, based on the dominant type signature observed in the subgraph.
    \item \textbf{Relation Prediction (RP)} classifies a drug--disease pair into one of four categories: \texttt{indication}, \texttt{contraindication}, \texttt{off-label use}, or \texttt{none}. Positive pairs are drawn from subgraph triples, while \texttt{none} examples are sampled from drug--disease pairs that do not appear in the subgraph.
    \item \textbf{Two-hop Verification (R1)} and \textbf{Two-hop Selection (R2)} are constructed from 2-hop contexts of the form Drug A $\rightarrow$ Disease B and Disease B $\rightarrow$ Disease C. Positive instances are created such that the queried Drug A $\rightarrow$ Disease C edge already exists in the subgraph. Negative instances preserve the same 2-hop scaffold but select a Disease C such that the queried edge is unsupported. R1 labels are sampled to achieve an approximately balanced ($\approx 50\%$) ``Yes'' rate, and R2 uses the same underlying 2-hop contexts.
    \item \textbf{Evidence-augmented Two-hop Verification (R1+E)} and \textbf{Evidence-augmented Two-hop Selection (R2+E)} extend the corresponding two-hop tasks by attaching short \kg{} feature-table snippets for the involved entities. Controlled sanitization is applied to redact lexical forms overlapping with relation answer options, thereby reducing potential answer leakage.
\end{itemize}

The ground-truth answers for these questions are strictly defined by triples in the extracted \kg{} mental-health subgraph. Negative options are constructed in a task-specific but KG-consistent manner: FC negatives are created by type-matched head or tail replacement under the same relation and retained only when the perturbed triple is unsupported by the extracted subgraph; RP uses \texttt{none} examples drawn from drug--disease pairs that do not appear in the subgraph; and negative R1/R2 instances preserve the same 2-hop scaffold but query a drug--disease edge that is unsupported by the subgraph. The final benchmark comprises 1{,}847 ET items, 2{,}000 EC items, 4{,}000 FC items, 7 RT items, 1{,}634 RP items, and 1{,}200 items each for R1, R1+E, R2, and R2+E.

% =========================================================
\section{Experiments}
\label{sec:exp}

\subsection{Models}
\label{subsec:models}

We evaluate 15 models spanning both closed- and open-source families: GPT-4.1, GPT-5.2-chat, GPT-4o, GPT-5-mini, GPT-5.1-chat, Qwen2.5-32B-Instruct, Mistral-7B-Instruct-v0.3, Qwen2.5-7B-Instruct, BioMistral-7B, Llama3-Med42-8B, DeepSeek-R1-Distill-Qwen-7B, DeepSeek-R1-Distill-Qwen-32B, Llama3.1-8B-Instruct, Meditron-7B, and Llama3-OpenBioLLM-8B. This set includes frontier GPT-series models, general-purpose open-source instruction-tuned models, and biomedical-domain variants. Representative technical reports and model cards for several evaluated families include GPT-4.1~\cite{openai2025gpt41}, GPT-5-mini~\cite{openai2026gpt5mini}, GPT-5.1-chat~\cite{openai2026gpt51chat}, GPT-5.2-chat~\cite{openai2026gpt52chat}, GPT-4o~\cite{hurst2024gpt}, Qwen2.5~\cite{qwen2024qwen25}, Mistral-7B-Instruct-v0.3~\cite{mistral7b_instruct_v03_hf}, BioMistral~\cite{labrak2024biomistral}, Med42~\cite{christophe2024med42}, DeepSeek-R1~\cite{guo2025deepseek}, Meditron~\cite{chen2023meditron}, OpenBioLLM~\cite{pal2024openbiollms}, and Llama 3~\cite{grattafiori2024llama}.

\subsection{Evaluation Protocol}
\label{subsec:protocol}

All tasks are evaluated under the same letter-only multiple-choice interface. For API-based models, we instruct the model to return a single option letter and apply strict answer parsing to recover one valid choice from the response. For local models, we use the same option-letter scoring setup as in the rest of the evaluation pipeline. Binary tasks use A/B labels rather than literal Yes/No to reduce lexical answer bias.

Because benchmark scoring under this setup depends on recovering a valid option letter, response validity is itself part of the evaluation problem in addition to raw task accuracy. We therefore treat output-format reliability as an important evaluation caveat in constrained multiple-choice assessment. Additional implementation details for benchmark construction, evidence sanitization, forced-choice local evaluation, API answer parsing, and randomness control are provided in Appendix~\ref{app:benchmark_inference_details}.

\subsection{Metrics}
\label{subsec:metrics}

We report task-level accuracy (\%) and grouped averages for four benchmark dimensions:
\begin{equation}
\mathrm{Avg}_E=\mathrm{mean}(\mathrm{ET},\mathrm{EC}),
\end{equation}
\begin{equation}
\mathrm{Avg}_R=\mathrm{mean}(\mathrm{FC},\mathrm{RT},\mathrm{RP}),
\end{equation}
\begin{equation}
\mathrm{Avg}_R^{\ast}=\mathrm{mean}(\mathrm{FC},\mathrm{RP}),
\end{equation}
\begin{equation}
\mathrm{Avg}_S=\mathrm{mean}(\mathrm{R1},\mathrm{R2}),
\end{equation}
\begin{equation}
\mathrm{Avg}_{S+E}=\mathrm{mean}(\mathrm{R1{+}E},\mathrm{R2{+}E}),
\end{equation}
\begin{equation}
\mathrm{Avg}_{All}=\mathrm{mean}\ \text{over all nine tasks},
\end{equation}
\begin{equation}
\mathrm{Avg}_{All}^{\ast}=\mathrm{mean}\ \text{over the eight tasks excluding RT}.
\end{equation}

Because RT contains only one question per retained relation (7 items total), its score should be interpreted cautiously. In the results discussion below, we therefore emphasize the starred averages when drawing overall comparisons that are less influenced by the small RT set.

Beyond task accuracy, we also report graph-oriented coverage over entities, relations, and triples in the curated mental-health slice of \kg{}. Specifically, we compute mean and degree-weighted correctness over entities and relations, together with a triple-level aggregate derived from entity and relation correctness. The \texttt{none} option in RP is treated as a task-specific no-relation label rather than as a KG relation and is therefore excluded from relation coverage. Full metric definitions are provided in Appendix~\ref{app:coverage_metric_defs}. In the current benchmark, all 1{,}847 entities and all 7 retained relations are measured for every model.

% =========================================================
\begin{table*}[t]
\caption{Benchmark accuracy (\%) on the \kg{} mental-health KG-to-QA tasks. We group tasks into four levels and report block averages: $\mathrm{Avg}_E{=}\mathrm{mean}(\mathrm{ET},\mathrm{EC})$, $\mathrm{Avg}_R{=}\mathrm{mean}(\mathrm{FC},\mathrm{RT},\mathrm{RP})$, $\mathrm{Avg}_R^{\ast}{=}\mathrm{mean}(\mathrm{FC},\mathrm{RP})$, $\mathrm{Avg}_S{=}\mathrm{mean}(\mathrm{R1},\mathrm{R2})$, $\mathrm{Avg}_{S+E}{=}\mathrm{mean}(\mathrm{R1{+}E},\mathrm{R2{+}E})$, $\mathrm{Avg}_{All}{=}\mathrm{mean}$ over all nine tasks, and $\mathrm{Avg}_{All}^{\ast}{=}\mathrm{mean}$ over the eight tasks excluding RT. Because RT contains only 7 items, the starred summaries are often more informative for overall comparisons. Best values in each column are bolded; ties are jointly bolded.}
\label{tab:main_results}
\centering
\footnotesize
\setlength{\tabcolsep}{2.2pt}
\renewcommand{\arraystretch}{1.08}
\begin{threeparttable}
\resizebox{\textwidth}{!}{%
\begin{tabular}{l|rrr|rrrrr|rrr|rrr|rr}
\toprule
\multirow{2}{*}{\textbf{Model}} &
\multicolumn{3}{c|}{\textbf{Entity}} &
\multicolumn{5}{c|}{\textbf{Relation}} &
\multicolumn{3}{c|}{\textbf{Subgraph}} &
\multicolumn{3}{c|}{\textbf{Evidence}} &
\multicolumn{2}{c}{\textbf{Overall}} \\
& \textbf{ET} & \textbf{EC} & \textbf{Avg$_E$} &
  \textbf{FC} & \textbf{RT (n=7)} & \textbf{RP} & \textbf{Avg$_R$} & \textbf{Avg$_{R^\ast}$} &
  \textbf{R1} & \textbf{R2} & \textbf{Avg$_S$} &
  \textbf{R1+E} & \textbf{R2+E} & \textbf{Avg$_{S+E}$} &
  \textbf{Avg$_{All}$} & \textbf{Avg$_{All^\ast}$}
\\
\midrule
GPT-4.1              & 97.62 & 91.85 & 94.73 & 63.32 & \textbf{100.00} & 54.96 & 72.76 & 59.14 & 57.58 & 64.00 & \textbf{60.79} & \textbf{71.42} & 61.50 & \textbf{66.46} & \textbf{73.58} & \textbf{70.28} \\
GPT-5.2              & 98.05 & 90.10 & 94.07 & 63.35 & \textbf{100.00} & \textbf{58.63} & \textbf{73.99} & \textbf{60.99} & 50.17 & \textbf{65.58} & 57.88 & 61.08 & \textbf{67.58} & 64.33 & 72.73 & 69.32 \\
GPT-4o               & 97.40 & 91.85 & 94.62 & 63.45 & \textbf{100.00} & 53.55 & 72.33 & 58.50 & \textbf{62.08} & 54.25 & 58.16 & 68.83 & 61.42 & 65.12 & 72.54 & 69.10 \\
GPT-5-mini           & \textbf{98.48} & 91.75 & 95.12 & \textbf{64.35} & \textbf{100.00} & 57.28 & 73.88 & 60.81 & 50.25 & 59.83 & 55.04 & 59.67 & 65.42 & 62.55 & 71.89 & 68.38 \\
GPT-5.1              & 98.27 & \textbf{92.35} & \textbf{95.31} & 62.50 & \textbf{100.00} & 58.08 & 73.53 & 60.29 & 50.17 & 60.83 & 55.50 & 60.00 & 64.42 & 62.21 & 71.85 & 68.33 \\
Qwen2.5-32B          & 76.45 & 54.60 & 65.53 & 58.40 & \textbf{100.00} & 38.43 & 65.61 & 48.41 & 50.50 & 59.00 & 54.75 & 61.25 & 50.08 & 55.66 & 60.97 & 56.09 \\
Mistral-7B           & 70.76 & 28.15 & 49.45 & 52.25 & 85.71  & 25.70 & 54.55 & 38.98 & 49.33 & 26.17 & 37.75 & 56.17 & 28.25 & 42.21 & 46.94 & 42.10 \\
Qwen2.5-7B           & 49.32 & 36.95 & 43.14 & 53.18 & 57.14  & 25.89 & 45.40 & 39.53 & 50.08 & 38.92 & 44.50 & 56.75 & 37.75 & 47.25 & 45.11 & 43.61 \\
BioMistral           & 29.24 & 14.80 & 22.02 & 49.90 & 28.57  & 20.99 & 33.15 & 35.45 & 53.92 & 22.92 & 38.42 & 53.17 & 38.42 & 45.80 & 34.66 & 35.42 \\
Med42-8B             & 24.96 & 24.55 & 24.76 & 50.30 & 14.29  & 27.97 & 30.85 & 39.13 & 55.00 & 24.92 & 39.96 & 51.42 & 36.17 & 43.80 & 34.40 & 36.91 \\
DeepSeek-R1-DQ-7B    &  9.42 & 20.40 & 14.91 & 49.60 & 28.57  & 32.62 & 36.93 & 41.11 & 49.75 & 27.17 & 38.46 & 49.75 & 25.00 & 37.38 & 32.48 & 32.96 \\
DeepSeek-R1-DQ-32B   &  6.55 & 18.90 & 12.72 & 48.15 & 14.29  & 22.46 & 28.30 & 35.30 & 49.75 & 33.92 & 41.84 & 49.58 & 35.75 & 42.66 & 31.04 & 33.13 \\
Llama3.1-8B          &  8.39 & 17.50 & 12.95 & 45.32 & 14.29  & 23.50 & 27.70 & 34.41 & 50.42 & 33.83 & 42.12 & 50.25 & 31.08 & 40.66 & 30.51 & 32.54 \\
Meditron             & 20.25 & 18.00 & 19.12 & 50.00 & 14.29  & 25.34 & 29.88 & 37.67 & 49.17 & 21.42 & 35.30 & 51.00 & 22.08 & 36.54 & 30.17 & 32.16 \\
OpenBioLLM-8B        & 16.57 & 23.35 & 19.96 & 50.25 &  0.00  & 23.68 & 24.64 & 36.97 & 49.75 & 24.58 & 37.16 & 56.50 & 24.33 & 40.41 & 29.89 & 33.63 \\
\bottomrule
\end{tabular}%
}
\begin{tablenotes}\scriptsize
\item \textbf{Model abbreviations:}
GPT-5.2=GPT-5.2-chat; GPT-5.1=GPT-5.1-chat; GPT-4o=GPT-4o; GPT-5-mini=GPT-5-mini;\\
Qwen2.5-32B=Qwen2.5-32B-Instruct; Qwen2.5-7B=Qwen2.5-7B-Instruct; Mistral-7B=Mistral-7B-Instruct-v0.3;\\
BioMistral=BioMistral-7B; Med42-8B=Llama3-Med42-8B; DeepSeek-R1-DQ-7B/32B=DeepSeek-R1-Distill-Qwen-7B/32B;\\
Llama3.1-8B=Llama3.1-8B-Instruct; Meditron=Meditron-7B; OpenBioLLM-8B=Llama3-OpenBioLLM-8B.
\end{tablenotes}
\end{threeparttable}
\end{table*}

\section{Results}

\subsection{Overall Performance}
\label{subsec:overall_results}

Table~\ref{tab:main_results} reports the accuracy of the 15 selected LLMs on MHGraphBench. Because RT contains only 7 items, we focus first on the RT-excluded overall summary $\mathrm{Avg}_{All}^{\ast}$. Under this summary, the strongest models in this evaluation are all GPT-series models: GPT-4.1 ranks first with $\mathrm{Avg}_{All}^{\ast}=70.28\%$, followed by GPT-5.2-chat at 69.32\% and GPT-4o at 69.10\%. GPT-5-mini and GPT-5.1-chat follow closely at 68.38\% and 68.33\%, respectively. The RT-including summary $\mathrm{Avg}_{All}$ yields a similar top-level ordering, with GPT-4.1 achieving the highest score at 73.58\%, followed by GPT-5.2-chat at 72.73\% and GPT-4o at 72.54\%.

Among open-source models, Qwen2.5-32B-Instruct is the strongest under both overall summaries, reaching 56.09\% on $\mathrm{Avg}_{All}^{\ast}$ and 60.97\% on $\mathrm{Avg}_{All}$. Under the RT-excluded summary, this still leaves a gap of more than 12 percentage points relative to the leading GPT models. Below Qwen2.5-32B-Instruct, performance drops markedly: Mistral-7B-Instruct-v0.3 reaches 42.10\% on $\mathrm{Avg}_{All}^{\ast}$ and Qwen2.5-7B-Instruct reaches 43.61\%. The remaining models cluster in the low- to mid-30s on the RT-excluded overall summary. Taken together, these results suggest that the benchmark is challenging not only for smaller biomedical models, but for most open-source models in general.

\subsection{Recognition vs.\ Relation Judgment}
\label{subsec:recognition_vs_judgment}

A central pattern in Table~\ref{tab:main_results} is the gap between recognition-oriented tasks and relation-judgment tasks. For the top GPT-series models, recognition-oriented performance is very strong. GPT-5.1-chat achieves the highest $\mathrm{Avg}_E$ at 95.31\%, closely followed by GPT-5-mini at 95.12\%, GPT-4.1 at 94.73\%, GPT-4o at 94.62\%, and GPT-5.2-chat at 94.07\%. ET is particularly strong, ranging from 97.40\% to 98.48\% across the top five models, while EC ranges from 90.10\% to 92.35\%. The RT subset is also saturated at 100.00\% for these models, but this result should be interpreted cautiously because RT contains only 7 items and is therefore better treated as a small descriptive subset than as strong standalone evidence.

However, this recognition strength does not translate into equally strong performance on judgment-related tasks. The best RP score is only 58.63\%, achieved by GPT-5.2-chat, followed by 58.08\% for GPT-5.1-chat and 57.28\% for GPT-5-mini. Even for the strongest models, these values remain far below ET and EC. The same separation appears in the grouped relation summaries. On the RT-excluded summary, GPT-5.2-chat reaches the highest $\mathrm{Avg}_R^{\ast}$ at 60.99\%, followed by GPT-5-mini at 60.81\% and GPT-5.1-chat at 60.29\%. The RT-including summary $\mathrm{Avg}_R$ shows a similar ordering, but it should be interpreted more cautiously because it includes the 7-item RT subset.

This pattern indicates that a model may correctly identify entity types and relation schemas while still struggling to distinguish whether a drug is indicated, contraindicated, used off-label, or unsupported for a target disorder.

\subsection{Short-Chain Reasoning}
\label{subsec:reasoning_results}

Short disease-mediated reasoning is one of the task categories that most clearly separates stronger models from weaker ones in the benchmark. Even among the strongest models, subgraph reasoning scores remain well below entity-level performance. GPT-4o achieves the highest R1 score at 62.08\%, while GPT-5.2-chat achieves the highest R2 score at 65.58\%. The grouped reasoning score $\mathrm{Avg}_S$ is highest for GPT-4.1 at 60.79\%, followed by GPT-4o at 58.16\% and GPT-5.2-chat at 57.88\%.

These values are notable because the reasoning tasks are tightly controlled: the 2-hop scaffold is explicitly provided, the answer space is constrained, and correctness is defined with respect to KG support. Even under these conditions, short-path composition remains substantially harder than ET or the small RT set. Among open-source models, the drop is sharper: Qwen2.5-32B-Instruct reaches 54.75\% on $\mathrm{Avg}_S$, whereas most others remain in the high-30s to mid-40s. This suggests that composing even two simple KG hops into a correct structured decision remains a major failure mode in constrained evaluation settings.

\subsection{Evidence Augmentation Is Not Uniformly Helpful}
\label{subsec:evidence_results}

Evidence augmentation affects models differently rather than providing a consistent benefit. On the positive side, several strong models improve when short KG-derived feature snippets are added. GPT-4.1 improves from 57.58\% to 71.42\% on R1, and GPT-4o improves from 62.08\% to 68.83\%. On the selection side, GPT-5.2-chat improves from 65.58\% to 67.58\% on R2, and GPT-5-mini improves from 59.83\% to 65.42\%. In grouped terms, GPT-4.1 achieves the best evidence-augmented reasoning score, with $\mathrm{Avg}_{S+E}=66.46\%$, followed by GPT-4o at 65.12\% and GPT-5.2-chat at 64.33\%.

At the same time, evidence is not uniformly helpful across models. Qwen2.5-32B-Instruct, for example, improves strongly on R1, from 50.50\% to 61.25\%, but drops sharply on R2, from 59.00\% to 50.08\%, yielding only a modest evidence-grouped score of 55.66\%. Smaller models also show inconsistent behavior, with some improving on one evidence-augmented task while remaining weak or deteriorating on the other. These results suggest that evidence augmentation is better interpreted as a diagnostic probe of whether a model can integrate short structured cues than as a universally corrective prompting strategy.

\subsection{Response-Format Reliability}
\label{subsec:bias_and_format}

Response-format reliability is an important evaluation issue in MHGraphBench because all tasks use a constrained letter-only multiple-choice interface. In this setting, measured performance depends not only on whether a model knows the correct answer, but also on whether it can reliably return a single valid option letter that can be unambiguously parsed. This issue is especially relevant for API-based models, whose outputs may include extra explanation, multiple candidate letters, or other text that does not strictly follow the requested response format. As a result, benchmark accuracy can partially reflect output controllability in addition to underlying task knowledge, a broader concern that has also been noted in prior work on multiple-choice LLM evaluation~\cite{wang2024beyond, pezeshkpour2024large, zheng2023large}.

This format issue also affects how chance-like scores on binary tasks such as FC or R1, where the positive and negative labels are approximately balanced, should be interpreted. Several weaker models remain close to 50\% accuracy on FC or R1 while simultaneously performing poorly on ET, EC, or RP. Such behavior could reflect weak but genuine reasoning ability, but it may also arise in part from unstable constrained outputs, response biases, or instruction-following failures under the letter-only evaluation setup. For this reason, aggregate task accuracy alone can be misleading unless it is interpreted together with output-validity checks and inspection of prediction distributions.

\subsection{Model-Family Observations}
\label{subsec:model_family_obs}

The results also offer a cautious perspective on biomedical-domain models. In this evaluation, biomedical or medically branded open-source models do not consistently outperform general-purpose instruction-tuned alternatives. BioMistral-7B~\cite{labrak2024biomistral}, Llama3-Med42-8B~\cite{christophe2024med42}, Meditron-7B~\cite{chen2023meditron}, and Llama3-OpenBioLLM-8B~\cite{pal2024openbiollms} all score below the strongest GPT models and below Qwen2.5-32B-Instruct on the overall summaries. Some of these models also exhibit unexpectedly weak entity-level performance: for example, Llama3-Med42-8B reaches only 24.76\% on $\mathrm{Avg}_E$, and Llama3-OpenBioLLM-8B reaches 19.96\%.

At the same time, these models are not uniformly weak across every dimension. Llama3-Med42-8B reaches 55.00\% on R1, and BioMistral-7B reaches 53.92\% on R1, despite their low entity scores. This uneven profile suggests that biomedical adaptation alone does not guarantee robust structured evaluation performance. However, this comparison should be interpreted cautiously, because the evaluated models also differ in parameter scale, base-model capability, instruction tuning, and output-format reliability. Taken together, the results suggest that performance in this benchmark reflects not only domain adaptation, but also the interaction among general model capacity, instruction following, constrained answer formats, and short reasoning requirements.

\begin{table}[t]
\caption{Compact knowledge coverage (\%) on the \kg{} mental-health subgraph. We report mean entity coverage, degree-weighted relation coverage, and triple coverage; full coverage metrics are provided in Appendix~\ref{app:coverage_results}. Models are sorted by $\mathrm{Cov}(T)$. Model abbreviations follow Table~\ref{tab:main_results}.}
\label{tab:coverage_results_main}
\centering
\scriptsize
\setlength{\tabcolsep}{3pt}
\resizebox{\columnwidth}{!}{%
\begin{tabular}{lrrr}
\toprule
\textbf{Model} & $\mathrm{CovAvg}(E)$ & $\mathrm{CovDeg}(R)$ & $\mathrm{Cov}(T)$ \\
\midrule
GPT-5-mini         & 77.81 & 63.30 & \textbf{65.27} \\
GPT-4o             & 77.36 & 61.18 & 64.77 \\
GPT-4.1            & 77.91 & 61.24 & 63.57 \\
GPT-5.1            & 77.84 & 59.62 & 61.48 \\
GPT-5.2            & 63.92 & 44.56 & 54.97 \\
Mistral-7B         & 51.19 & 49.21 & 53.83 \\
Qwen2.5-7B         & 44.61 & 50.35 & 53.20 \\
Qwen2.5-32B        & 61.47 & 55.09 & 52.31 \\
Meditron           & 32.59 & 47.65 & 48.46 \\
DeepSeek-R1-DQ-7B  & 29.24 & 47.25 & 47.34 \\
DeepSeek-R1-DQ-32B & 26.54 & 45.77 & 44.31 \\
Med42-8B           & 38.23 & 48.34 & 38.89 \\
OpenBioLLM-8B      & 34.02 & 48.10 & 37.26 \\
BioMistral         & 36.74 & 46.34 & 37.12 \\
Llama3.1-8B        & 26.67 & 42.98 & 36.71 \\
\bottomrule
\end{tabular}%
}
\end{table}

\subsection{Knowledge Coverage}
\label{subsec:coverage_results}

Coverage provides a complementary graph-wide view of model performance (Table~\ref{tab:coverage_results_main}; full metrics in Appendix~\ref{app:coverage_results}). GPT-5-mini achieves the strongest triple coverage, with $\mathrm{Cov}(T)=65.27\%$, even though GPT-4.1 remains the top model by average task accuracy. This shows that benchmark averages and graph-wide coverage are not interchangeable. Coverage also changes the interpretation of open-source models: Qwen2.5-32B-Instruct is the best open-source model by $\mathrm{Avg}_{All}^{\ast}$, but not by triple coverage, and GPT-5.2-chat shows lower coverage than the other GPT models despite ranking near the top on the main task table.

\subsection{Refined Entity and Relation Analysis}
\label{subsec:fine_grained_analysis}

To better understand where models succeed or fail, we also compute fine-grained entity- and relation-centric accuracy, with full tables reported in Appendix~\ref{app:fine_grained_results}. The fine-grained relation results show that \texttt{contraindication} is by far the hardest retained relation on average, whereas \texttt{indication} is comparatively easier. The fine-grained entity results further show that high benchmark incidence does not guarantee ease: anxiety-spectrum and psychotic-spectrum entities remain difficult despite their prominence in the benchmark. Taken together, these results suggest that model failures are concentrated in clinically important and diagnostically heterogeneous parts of the graph.

% =========================================================
\section{Discussion and Conclusion}
\label{sec:discussion_conclusion}

We introduced MHGraphBench, a KG-grounded benchmark for evaluating mental-health biomedical knowledge in LLMs using a curated 1-hop mental-health subgraph of \kg{}. The benchmark transforms KG-backed facts into nine standardized multiple-choice task families spanning entity recognition, relation judgment, and short two-hop reasoning, and complements task accuracy with graph-wide coverage and fine-grained entity- and relation-centric analyses.

Across 15 models, our results reveal a persistent recognition-to-judgment gap. Leading models achieve near-perfect performance on entity typing and very strong performance on the small relation-typing subset, yet they remain substantially weaker on relation prediction and short-chain reasoning. Coverage analysis further shows that average task accuracy and graph-wide coverage are not interchangeable: GPT-4.1 performs best on the main benchmark averages, whereas GPT-5-mini achieves the strongest triple coverage. Fine-grained analyses localize especially difficult graph regions, with clinically important relations such as \texttt{contraindication} and prominent entities such as \emph{anxiety disorder} remaining challenging. We also find that evidence augmentation is not uniformly helpful across models and that response-format reliability can materially affect measured performance under constrained multiple-choice evaluation.

These findings suggest that broad biomedical competence should not be equated with reliable structured judgment in mental-health settings. They are consistent with recent evidence outside biomedicine. In a recent expert-led study on high-temperature superconductivity, LLM systems grounded in curated literature outperformed more general systems, yet all evaluated systems still showed important limitations in expert-level scientific question answering~\cite{guo2026expert}. Taken together, these results suggest that current LLMs may read and organize scientific text fluently without consistently supporting the deeper judgment required for expert reasoning.

More broadly, our results show that KG-grounded benchmarking provides an interpretable and reproducible way to study what LLMs capture about mental-health biomedical structure, while highlighting limitations in safety-relevant relation distinctions and controlled reasoning. Future work should move toward more challenging but still controlled benchmarks that better connect structured knowledge evaluation with clinically relevant mental-health decision support. MHGraphBench should therefore be interpreted as a structured evaluation of a curated KG slice rather than as a direct assessment of real-world clinical safety. % the benchmark should be interpreted as a structured evaluation with respect to a curated KG slice rather than as a direct assessment of real-world clinical safety.

% =========================================================
\section*{Limitations}

Our benchmark inherits the coverage limits and curation decisions of \kg{} as well as those of our mental-health subgraph extraction process. As a result, the task suite is intentionally scoped and does not capture the full breadth of psychiatric care, longitudinal patient context, or individualized treatment decision-making.

All labels are defined with respect to the extracted \kg{} mental-health subgraph. Because biomedical knowledge and clinical guidelines evolve over time, these KG-based labels may be incomplete or may lag behind the most up-to-date evidence. Accordingly, the benchmark measures agreement with a curated KG slice rather than absolute clinical truth.

In addition, we do not perform manual or expert validation of sampled benchmark items, negative instances, or evidence snippets beyond the KG-grounded construction pipeline itself. This means that benchmark validity depends on the quality of the underlying graph, the extraction procedure, and the task-generation rules. In particular, an edge being unsupported in the extracted subgraph should not be interpreted as evidence that the corresponding claim is false in the real world; it indicates only that the queried relation is absent from the curated benchmark graph. Similarly, although evidence snippets are sanitized to reduce direct answer leakage, they are not externally adjudicated by domain experts for completeness, clinical appropriateness, or real-world decision support value.

Although coverage and fine-grained analyses provide a richer view than average task accuracy alone, they still depend on benchmark construction choices and on how task items involve particular graph components. These analyses help localize strengths and weaknesses, but they should not be interpreted as exhaustive measurements of mental-health biomedical knowledge.

Finally, because evaluation relies on constrained multiple-choice outputs, models that fail to follow answer-format instructions may be penalized for reasons partly independent of their underlying biomedical reasoning ability. This is both a limitation and an empirical finding of the benchmark: output controllability is entangled with measured performance.

\section*{Ethics Statement}

This work does not evaluate clinical safety, real-world mental-health decision-making, or patient-specific treatment appropriateness. The benchmark should not be used as a substitute for expert oversight, especially in settings involving treatment boundaries, contraindications, or crisis-related decisions~\cite{agarwal2024medhalu, zhu2025can}. More broadly, our results should be interpreted as a structured evaluation of KG-grounded mental-health biomedical knowledge with respect to a curated mental-health subgraph, rather than as evidence of clinical validity, real-world safety, or readiness for clinical deployment.

\section*{Acknowledgments}

This research was, in part, funded by the Advanced Research Projects Agency for Health (ARPA-H). The views and conclusions contained in this document are those of the authors and should not be interpreted as representing the official policies, either expressed or implied, of the United States Government.
% =========================================================
\bibliography{refs}

% =========================================================
\appendix

\section{Subgraph Statistics}
\label{app:subgraph_stats}

This appendix section summarizes the size and composition of the curated \kg{} mental-health subgraph used throughout the benchmark.

\subsection{Summary Statistics}

\begin{table}[t]
\caption{Summary statistics of the \kg{} mental-health subgraph used in this study. 
HP = high-precision; canon./dedup. = canonicalization and deduplication.}
\label{tab:subgraph_stats_summary}
\centering
\footnotesize
\setlength{\tabcolsep}{4pt}
\begin{tabular}{@{}lr@{}}
\toprule
\textbf{Item} & \textbf{Count} \\
\midrule
HP candidate disease nodes & 44 \\
Post-curation exclusions & 2 \\
Final seed disease nodes & 42 \\
Raw seed-touching edges & 9{,}242 \\
Canon./dedup. unique triples & 4{,}621 \\
Unique entities & 1{,}847 \\
Retained relation types & 7 \\
\bottomrule
\end{tabular}
\end{table}

\subsection{Entity and Relation Breakdown}

\begin{table}[t]
\caption{Entity-type and relation-type counts in the \kg{} mental-health subgraph.}
\label{tab:subgraph_stats_breakdown}
\centering
\footnotesize
\begin{tabular}{lr}
\toprule
\textbf{Entity type} & \textbf{Count} \\
\midrule
Disease & 75 \\
Drug & 345 \\
Gene/protein & 1{,}326 \\
Effect/phenotype & 66 \\
Exposure & 35 \\
\midrule
\textbf{Relation type} & \textbf{Count} \\
\midrule
\texttt{disease\_protein} & 3{,}565 \\
\texttt{contraindication} & 556 \\
\texttt{indication} & 234 \\
\texttt{disease\_disease} & 84 \\
\texttt{disease\_phenotype\_positive} & 82 \\
\texttt{off-label use} & 54 \\
\texttt{exposure\_disease} & 46 \\
\bottomrule
\end{tabular}
\end{table}

\section{Mental-Health Seed Disease Nodes}
\label{app:seed_list}

This appendix section documents how the psychiatric seed disease list was defined and reports the final set of seed nodes used for subgraph extraction.

\subsection{Seed Selection Procedure}

We began with a manually curated high-precision candidate seed list of 44 PrimeKG disease nodes, stored in \texttt{mental\_health\_seed\_diseases\_HP.csv}. PrimeKG disease nodes are encoded using terms from the Mondo Disease Ontology (MONDO) and grouped into clinically meaningful disease nodes during PrimeKG construction~\cite{chandak2023building}. Candidate selection was restricted to psychiatric disorders and closely related conditions intended to define the benchmark scope. We then applied two manual post-curation exclusions: one out-of-scope entry (\emph{X-linked intellectual disability-psychosis-macroorchidism syndrome}), which appeared in the initial candidate list but was excluded at post-curation because it was outside the intended psychiatric benchmark scope, and one outdated entry (\emph{multiple personality disorder}). This yielded the final set of 42 psychiatric seed disease nodes used for subgraph extraction.

\subsection{Final Seed List}

The final seed list, stored in \texttt{mental\_health\_seed\_diseases\_FINAL.csv}, is shown in Table~\ref{tab:seed_disease_list}.

\begin{table*}[t]
\caption{Final set of 42 psychiatric seed disease nodes used to extract the \kg{} mental-health subgraph.}
\label{tab:seed_disease_list}
\centering
\scriptsize
\setlength{\tabcolsep}{4pt}
\renewcommand{\arraystretch}{1.05}
\begin{tabular}{r p{0.28\textwidth} r p{0.28\textwidth}}
\toprule
\textbf{Node index} & \textbf{Node name} & \textbf{Node index} & \textbf{Node name} \\
\midrule
27933 & anxiety disorder & 84190 & binge eating disorder \\
28249 & major affective disorder & 84195 & narcissistic personality disorder \\
28313 & schizophrenia & 84204 & mixed anxiety and depressive disorder \\
28592 & bulimia nervosa, susceptibility to & 84208 & alcoholic psychosis \\
28899 & obsessive-compulsive disorder & 84226 & postpartum depression \\
32965 & early-onset schizophrenia & 84288 & antisocial personality disorder \\
33572 & psychotic disorder & 84294 & paranoid schizophrenia \\
35758 & specific phobia & 94658 & neurotic depression \\
36021 & personality disorder & 95043 & avoidant personality disorder \\
38242 & bipolar disorder & 95044 & dependent personality disorder \\
38945 & bulimia nervosa & 95390 & schizotypal personality disorder \\
38957 & major depressive disorder & 95419 & schizoid personality disorder \\
39833 & agoraphobia & 95420 & paranoid personality disorder \\
83763 & manic bipolar affective disorder & 95557 & atypical depressive disorder \\
83779 & schizoaffective disorder & 95941 & histrionic personality disorder (disease) \\
83840 & unipolar depression & 96890 & substance-induced psychosis \\
83841 & endogenous depression & 97059 & treatment-refractory schizophrenia \\
83842 & anorexia nervosa & 97074 & methamphetamine-induced psychosis \\
83903 & post-traumatic stress disorder & 97811 & anorexia nervosa, susceptibility to, 1 \\
83904 & social phobia & 98548 & postpartum psychosis \\
83910 & drug psychosis & 99866 & panic disorder without or with agoraphobia \\
\bottomrule
\end{tabular}
\end{table*}

\section{Coverage Metric Definitions}
\label{app:coverage_metric_defs}

This section defines the coverage metrics used in the main paper. We first define per-entity and per-relation correctness and then derive entity-, relation-, and triple-level coverage scores.

\subsection{Per-Entity and Per-Relation Correctness}

Beyond task accuracy, we quantify how well a model covers the mental-health slice of \kg{} in terms of correctness over entities, relations, and triples. Let the curated mental-health graph be $G=(V,R,T)$, where $V$ is the entity set, $R$ is the relation set, and $T \subseteq V \times R \times V$ is the triple set.

For each entity $e\in V$, let $\mathcal{Q}(e)$ denote the set of benchmark items whose gold annotation involves $e$. We define empirical entity correctness as
\begin{equation}
a_E(e)=\frac{1}{|\mathcal{Q}(e)|}\sum_{q\in \mathcal{Q}(e)} \mathbf{1}[\hat{y}_q=y_q].
\end{equation}
Similarly, for each relation $r\in R$, let $\mathcal{Q}(r)$ denote the set of benchmark items whose gold annotation involves $r$, and define empirical relation correctness as
\begin{equation}
a_R(r)=\frac{1}{|\mathcal{Q}(r)|}\sum_{q\in \mathcal{Q}(r)} \mathbf{1}[\hat{y}_q=y_q].
\end{equation}

The \texttt{none} option is treated as a task-specific no-relation label rather than a KG relation. It is therefore excluded from relation coverage and contributes only indirectly through entity-level correctness.

\subsection{Coverage Scores}

We then define five coverage scores. Mean entity coverage is
\begin{equation}
\mathrm{CovAvg}(E)=\frac{1}{|V_{\mathrm{meas}}|}\sum_{e\in V_{\mathrm{meas}}} a_E(e).
\end{equation}
For degree-weighted entity coverage, we define entity degree as the number of incident triples in $T$,
\begin{equation}
\deg(e)=\left|\{(h,r,t)\in T : h=e \ \text{or} \ t=e\}\right|,
\end{equation}
with normalizing constant
\begin{equation}
Z_E=\sum_{e\in V}\deg(e).
\end{equation}
The resulting degree-weighted entity coverage is
\begin{equation}
\mathrm{CovDeg}(E)=\frac{1}{Z_E}\sum_{e\in V}\deg(e)\,a_E(e).
\end{equation}

For relations, mean relation coverage is
\begin{equation}
\mathrm{CovAvg}(R)=\frac{1}{|R_{\mathrm{meas}}|}\sum_{r\in R_{\mathrm{meas}}} a_R(r),
\end{equation}
where relation degree is defined as the number of triples in $T$ that use relation $r$,
\begin{equation}
\deg(r)=\left|\{(h,r',t)\in T : r'=r\}\right|.
\end{equation}
The corresponding degree-weighted relation coverage is
\begin{equation}
\mathrm{CovDeg}(R)=\frac{1}{|T|}\sum_{r\in R}\deg(r)\,a_R(r).
\end{equation}

Finally, for each triple $(h,r,t)\in T$, we define an auxiliary triple score
\begin{equation}
s(h,r,t)=\frac{a_E(h)+a_R(r)+a_E(t)}{3},
\end{equation}
and triple coverage as
\begin{equation}
\mathrm{Cov}(T)=\frac{1}{|T|}\sum_{(h,r,t)\in T} s(h,r,t).
\end{equation}

Here, $V_{\mathrm{meas}}$ and $R_{\mathrm{meas}}$ denote the sets of measured entities and relations. In the current benchmark, all 1{,}847 entities and all 7 retained relations are measured for every model.

\section{Benchmark Construction, Evidence, and Evaluation Details}
\label{app:benchmark_inference_details}

This appendix section provides implementation-level details that extend, rather than repeat, the benchmark overview in Section~\ref{sec:dataset_method} and the evaluation description in Section~\ref{subsec:protocol}. Unless otherwise noted, all statements in this section are derived directly from the benchmark-generation and evaluation code used in our experiments.

\subsection{Benchmark Construction Details}
\label{app:benchmark_construction_details}

The benchmark is generated from \kg{} using four input files: \texttt{kg.csv}, \texttt{mental\_health\_seed\_diseases\_FINAL.csv}, \texttt{disease\_features.tab}, \texttt{drug\_features.tab}. The final seed file contains the 42 psychiatric seed disease nodes reported in Appendix~\ref{app:seed_list}. Starting from these seeds, we extract all 1-hop seed-touching edges from \texttt{kg.csv}, retain only a fixed set of seven clinically salient relations, canonicalize relation direction to predefined head/tail type signatures, and deduplicate the resulting triples.

The retained relations are \texttt{disease\_protein}, \texttt{contraindication}, \texttt{indication}, \texttt{off-label use}, \texttt{disease\_disease}, \texttt{disease\_phenotype\_positive}, and \texttt{exposure\_disease}. Canonical relation signatures are fixed during preprocessing: \texttt{disease\_protein} is canonicalized as disease$\rightarrow$gene/protein; \texttt{contraindication}, \texttt{indication}, and \texttt{off-label use} as drug$\rightarrow$disease; \texttt{disease\_disease} as disease$\rightarrow$disease; \texttt{disease\_phenotype\_positive} as disease$\rightarrow$effect/phenotype; and \texttt{exposure\_disease} as exposure$\rightarrow$disease. For \texttt{disease\_disease}, symmetric duplicates are additionally removed by lexicographic canonicalization of the two disease names.

All task instances are generated programmatically from fixed English templates and use a unified letter-only answer interface. Entity Typing (ET) is a 5-way multiple-choice question over entity types. Entity Clustering (EC) is constructed as an odd-one-out task with four entities of one type and one entity of another type. Relation Typing (RT) asks for the dominant head$\rightarrow$tail type signature of a relation. Relation Prediction (RP) is a 4-way multiple-choice task over \texttt{indication}, \texttt{contraindication}, \texttt{off-label use}, and \texttt{none}. Two-hop Verification (R1) is a binary A/B task, and Two-hop Selection (R2) is a 4-way multiple-choice task. Evidence-augmented variants (R1+E and R2+E) use the same underlying task structure but append short feature-table evidence snippets to the question.

The two-hop tasks are constructed from contexts in which a drug is linked to disease A by one of the three drug--disease usage relations and disease A is linked to disease B by \texttt{disease\_disease}. Positive R1 instances are those for which the queried drug--disease B relation already exists in the retained subgraph. Negative R1 instances preserve the same 2-hop scaffold but require that the queried drug--disease B relation be absent from the subgraph. The generator targets an approximately balanced R1 label distribution with a 50\% Yes rate and does not allow the intermediate disease and queried disease to be identical. R2 instances are built from the same contexts and ask the model to select the most appropriate relation label for the queried drug--disease pair.

The Fact Checking (FC) task is balanced \emph{per relation}. For each retained relation \(r\), the benchmark generator samples positive triples from that relation and constructs an equal number of negative examples under the same relation. FC negatives are created by type-matched head or tail replacement while preserving the original relation label, and the perturbed triple is kept only if it does not appear in the retained mental-health subgraph. This design avoids relation-replacement negatives and makes per-relation FC behavior easier to interpret.

\subsection{Evidence Construction and Sanitization}
\label{app:evidence_sanitization_details}

Evidence-augmented tasks draw text snippets from \texttt{disease\_features.tab} and \texttt{drug\_features.tab}. For disease nodes, the pipeline attempts to use available fields such as \texttt{mondo\_name}, \texttt{mondo\_definition}, \texttt{umls\_description}, \texttt{orphanet\_clinical\_description}, \texttt{mayo\_symptoms}, \texttt{mayo\_causes}, \texttt{mayo\_risk\_factors}, and \texttt{orphanet\_management\_and\_treatment}. For drug nodes, it attempts to use fields such as \texttt{description}, \texttt{indication}, \texttt{mechanism\_of\_action}, \texttt{pharmacodynamics}, \texttt{half\_life}, \texttt{state}, and \texttt{category}. When multiple rows are available for the same node, the generator keeps the first non-empty value for each field. Long text fields are truncated to at most 220 characters per field before question assembly.

To reduce answer leakage, evidence text is sanitized before insertion into the question. The sanitization step redacts lexical forms overlapping with relation answer options, including patterns matching \texttt{indication}, \texttt{contraindication}, and \texttt{off-label}, and replaces them with a neutral placeholder \texttt{[REL]}. In addition, the original drug-table field name \texttt{indication} is rendered with the more neutral display label \emph{Clinical use} when evidence blocks are assembled. The evidence block is then attached in a fixed order: drug evidence first, followed by evidence for disease A and disease B.

\subsection{Evaluation Protocol Details}
\label{app:evaluation_protocol_details}

All benchmark tasks are evaluated under the same letter-only interface. For binary tasks, the benchmark uses A/B labels rather than literal Yes/No strings, with A corresponding to Yes and B corresponding to No. The evaluation code records per-task accuracy for all tasks and additionally computes diagnostic quantities such as prediction-\(A\) rate and balanced accuracy for A/B tasks.

For local Hugging Face models, evaluation is performed by forced-choice scoring rather than free-form generation. The prompt is constructed by appending the fixed anchor \texttt{\textbackslash nAnswer:} to each benchmark question. If the tokenizer provides a chat template, the prompt is wrapped using the tokenizer's chat-template interface; otherwise, the raw question text is used directly. Candidate answer letters are scored through multi-token log-probability accumulation over several surface forms, including \texttt{(A)}, bare-letter forms with a trailing newline, parenthesized-letter forms with a trailing newline, and several short prefixes. Scores for multiple surface forms corresponding to the same letter are merged by taking the maximum score for that letter, and the highest-scoring allowed option is selected as the model prediction. This procedure makes local evaluation deterministic given fixed model weights, tokenizer behavior, and benchmark inputs.

The local evaluation code uses batch size 1, maximum sequence length 4096, \texttt{bfloat16} model loading, and automatic device placement via \texttt{device\_map="auto"}. Model loading is performed with \texttt{trust\_remote\_code=True}. For API-based models, the evaluation pipeline queries the model with temperature set to 0 and a maximum completion length of 120 tokens, then applies strict answer parsing to recover a single option letter from the returned text. The parser first searches for explicit answer patterns such as \texttt{Answer: (X)} and then falls back to leading-letter or in-text letter matching when necessary. Because some API-based models occasionally return outputs that do not perfectly follow the requested constrained format, response validity is itself an important part of the measured evaluation behavior.

\subsection{Randomness and Deterministic Settings}
\label{app:randomness_details}

Benchmark generation uses a fixed Python random seed of 42. This seed controls the sampling operations used during task construction, including option shuffling, entity selection, negative sampling, and task-instance ordering. Local model evaluation by forced-choice scoring does not sample from the model and is therefore deterministic given fixed model weights, tokenizer behavior, and benchmark inputs. API-based evaluation is configured with temperature 0 to reduce generation variability.

\section{Extended Coverage and Fine-Grained Results}
\label{app:extended_results}

This appendix section reports the full coverage tables and the detailed entity- and relation-level analyses that complement the compact summaries in the main text.

\subsection{Knowledge Coverage Results}
\label{app:coverage_results}

\begin{table*}[t]
\caption{Knowledge coverage (\%) on the \kg{} mental-health subgraph. Higher is better. All models measure all 1{,}847 entities and all 7 retained relations. Models are sorted by $\mathrm{Cov}(T)$.}
\label{tab:coverage_results}
\centering
\footnotesize
\setlength{\tabcolsep}{3.2pt}
\begin{tabular}{lrrrrrrr}
\toprule
\textbf{Model} & $\mathrm{CovAvg}(E)$ & $\mathrm{CovDeg}(E)$ & $\mathrm{CovAvg}(R)$ & $\mathrm{CovDeg}(R)$ & $\mathrm{Cov}(T)$ & \textbf{Meas.\ E} & \textbf{Meas.\ R} \\
\midrule
GPT-5-mini                   & 77.81 & 66.26 & 64.34 & 63.30 & \textbf{65.27} & 1847 & 7 \\
GPT-4o                       & 77.36 & 66.57 & 61.23 & 61.18 & 64.77 & 1847 & 7 \\
GPT-4.1                      & 77.91 & 64.74 & 61.33 & 61.24 & 63.57 & 1847 & 7 \\
GPT-5.1-chat                 & 77.84 & 62.40 & 62.30 & 59.62 & 61.48 & 1847 & 7 \\
GPT-5.2-chat                 & 63.92 & 60.17 & 44.86 & 44.56 & 54.97 & 1847 & 7 \\
Mistral-7B-Instruct-v0.3     & 51.19 & 56.14 & 52.52 & 49.21 & 53.83 & 1847 & 7 \\
Qwen2.5-7B-Instruct          & 44.61 & 54.63 & 54.92 & 50.35 & 53.20 & 1847 & 7 \\
Qwen2.5-32B-Instruct         & 61.47 & 50.92 & 61.62 & 55.09 & 52.31 & 1847 & 7 \\
Meditron-7B                  & 32.59 & 48.87 & 44.68 & 47.65 & 48.46 & 1847 & 7 \\
DeepSeek-R1-Distill-Qwen-7B  & 29.24 & 47.39 & 44.89 & 47.25 & 47.34 & 1847 & 7 \\
DeepSeek-R1-Distill-Qwen-32B & 26.54 & 43.58 & 46.11 & 45.77 & 44.31 & 1847 & 7 \\
Llama3-Med42-8B              & 38.23 & 34.17 & 46.46 & 48.34 & 38.89 & 1847 & 7 \\
Llama3-OpenBioLLM-8B         & 34.02 & 31.84 & 45.81 & 48.10 & 37.26 & 1847 & 7 \\
BioMistral-7B                & 36.74 & 32.50 & 45.84 & 46.34 & 37.12 & 1847 & 7 \\
Llama3.1-8B-Instruct         & 26.67 & 33.58 & 44.39 & 42.98 & 36.71 & 1847 & 7 \\
\bottomrule
\end{tabular}
\end{table*}

Table~\ref{tab:coverage_results} provides a complementary graph-wide view of performance. Because all 1{,}847 entities and all 7 retained relations are measured for every model, these metrics summarize behavior over the full benchmark graph rather than over a partial subset. The strongest triple coverage is achieved by GPT-5-mini, with $\mathrm{Cov}(T)=65.27\%$, followed by GPT-4o at 64.77\% and GPT-4.1 at 63.57\%. This ranking differs from the main task average, where GPT-4.1 is the top model. The discrepancy suggests that average task accuracy and graph-wide coverage capture different aspects of model behavior.

At the entity level, GPT-4.1 achieves the highest mean entity coverage, with $\mathrm{CovAvg}(E)=77.91\%$, closely followed by GPT-5.1-chat and GPT-5-mini. However, GPT-5-mini attains the highest degree-weighted relation coverage, with $\mathrm{CovDeg}(R)=63.30\%$, which helps explain why it leads on triple coverage. In other words, GPT-5-mini is especially strong on graph-central relation mass, even though GPT-4.1 remains slightly stronger on average benchmark accuracy.

Coverage also changes the interpretation of open-source models. Qwen2.5-32B-Instruct is the best open-source model by $\mathrm{Avg}_{All}$ in Table~\ref{tab:main_results}, but it does not have the strongest open-source triple coverage. Mistral-7B-Instruct-v0.3 and Qwen2.5-7B-Instruct both slightly exceed Qwen2.5-32B-Instruct on $\mathrm{Cov}(T)$, suggesting that correctness on high-degree graph components can differ from overall task-level performance. GPT-5.2-chat is another notable case: despite ranking near the top on the main benchmark averages, its coverage scores are substantially lower than those of the other GPT models, indicating that its correctness is less evenly distributed across the graph.

\subsection{Fine-Grained Entity and Relation Analysis}
\label{app:fine_grained_results}

For a model $m$ and relation $r$, we define
\begin{equation}
\mathrm{Acc}_m(r)=\frac{c_m(r)}{n(r)},
\end{equation}
where $n(r)$ is the number of benchmark items whose gold annotation involves relation $r$, and $c_m(r)$ is the number of those items answered correctly. Similarly, for an entity $e$, we define
\begin{equation}
\mathrm{Acc}_m(e)=\frac{c_m(e)}{n(e)}.
\end{equation}
We report high-incidence entities and retained relations because these components exert a strong influence on observed benchmark behavior and help localize where model performance concentrates.

\begin{table*}[t]
\caption{Fine-grained accuracy (\%) on retained relations in the mental-health subgraph. Relations are ordered by benchmark incidence in the evaluation set. The highest model score in each row is bolded. The final column reports the mean accuracy across models.}
\label{tab:fine_grained_relations}
\centering
\scriptsize
\setlength{\tabcolsep}{2.1pt}
\renewcommand{\arraystretch}{1.04}
\resizebox{\textwidth}{!}{%
\begin{threeparttable}
\begin{tabular}{l r | r r r r r r r r r r r r r r r | r}
\toprule
\textbf{Relation} & \textbf{Items} &
\textbf{BioM7B} & \textbf{DS32} & \textbf{DS7} & \textbf{Med42} & \textbf{OpenBio} &
\textbf{Meditron} & \textbf{Mistral7B} & \textbf{Q32} & \textbf{Q7} &
\textbf{GPT4.1} & \textbf{GPT4o} & \textbf{GPT5m} & \textbf{GPT5.1} & \textbf{GPT5.2} & \textbf{L3.1-8B} &
\textbf{Mean} \\
\midrule
contraindication             & 3348 & 24.5 & 38.1 & 35.7 & 39.8 & 38.9 & 36.8 & 27.0 & 33.1 & 30.3 & \textbf{43.6} & 41.8 & 37.8 & 38.0 & 31.8 & 39.8 & 35.8 \\
disease\_protein             & 2983 & 49.5 & 47.4 & 49.7 & 49.9 & 50.1 & 50.0 & 50.6 & 56.1 & 51.0 & 62.9 & 62.9 & \textbf{65.9} & 61.4 & 46.0 & 43.3 & 53.1 \\
indication                   & 1629 & 45.0 & 32.7 & 35.8 & 45.0 & 39.7 & 38.0 & 75.3 & 80.4 & \textbf{83.9} & 78.1 & 81.9 & 81.7 & 77.9 & 51.9 & 38.6 & 59.1 \\
off-label use                & 239  & 50.5 & 49.5 & 41.7 & 41.7 & 41.7 & 39.8 & 40.8 & 66.0 & 49.5 & 63.1 & 62.1 & 67.0 & \textbf{70.9} & 49.5 & 41.7 & 51.7 \\
disease\_disease             & 95   & 50.5 & 48.4 & 49.5 & 50.5 & 51.6 & 49.5 & 65.3 & \textbf{72.6} & 55.8 & 63.2 & 64.2 & 64.2 & 69.5 & 44.2 & 49.5 & 56.6 \\
disease\_phenotype\_positive & 93   & 51.6 & 52.7 & 49.5 & 50.5 & 49.5 & 49.5 & 51.6 & \textbf{61.3} & 47.3 & 58.1 & 57.0 & 59.1 & 58.1 & 39.8 & 54.8 & 52.7 \\
exposure\_disease            & 63   & 49.2 & 54.0 & 52.4 & 47.6 & 49.2 & 49.2 & 57.1 & 61.9 & 66.7 & 60.3 & 58.7 & \textbf{74.6} & 60.3 & 50.8 & 42.9 & 55.7 \\
\bottomrule
\end{tabular}
\begin{tablenotes}\scriptsize
\item \textbf{Model abbreviations:}
BioM7B=BioMistral-7B;
DS32=DeepSeek-R1-Distill-Qwen-32B;
DS7=DeepSeek-R1-Distill-Qwen-7B;
Med42=Llama3-Med42-8B;
OpenBio=Llama3-OpenBioLLM-8B;
Meditron=Meditron-7B;
Mistral7B=Mistral-7B-Instruct-v0.3;
Q32=Qwen2.5-32B-Instruct;
Q7=Qwen2.5-7B-Instruct;
GPT4.1=GPT-4.1;
GPT4o=GPT-4o;
GPT5m=GPT-5-mini;
GPT5.1=GPT-5.1-chat;
GPT5.2=GPT-5.2-chat;
L3.1-8B=Llama3.1-8B-Instruct.
\end{tablenotes}
\end{threeparttable}
}
\end{table*}

\begin{table*}[t]
\caption{Fine-grained accuracy (\%) on Top-15 high-incidence mental-health entities in the benchmark. Entities are ordered by benchmark incidence in the evaluation set. The final column reports the mean accuracy across models.}
\label{tab:fine_grained_entities}
\centering
\scriptsize
\setlength{\tabcolsep}{2.0pt}
\renewcommand{\arraystretch}{1.04}
\resizebox{\textwidth}{!}{%
\begin{threeparttable}
\begin{tabular}{l r | r r r r r r r r r r r r r r r | r}
\toprule
\textbf{Entity} & \textbf{Items} &
\textbf{BioM7B} & \textbf{DS32} & \textbf{DS7} & \textbf{Med42} & \textbf{OpenBio} &
\textbf{Meditron} & \textbf{Mistral7B} & \textbf{Q32} & \textbf{Q7} &
\textbf{GPT4.1} & \textbf{GPT4o} & \textbf{GPT5m} & \textbf{GPT5.1} & \textbf{GPT5.2} & \textbf{L3.1-8B} &
\textbf{Mean} \\
\midrule
anxiety disorder                          & 1178 & 24.8 & 57.0 & 53.9 & 28.6 & 28.3 & 57.7 & 46.4 & 30.0 & 54.2 & 36.5 & 41.4 & 38.3 & 30.3 & 45.7 & 28.7 & 40.1 \\
bipolar disorder                          & 758  & 28.5 & 34.6 & 61.7 & 30.7 & 30.2 & 62.0 & 60.2 & 40.0 & 62.2 & 59.5 & 59.8 & 57.6 & 51.7 & 61.2 & 31.7 & 48.8 \\
major affective disorder                  & 681  & 28.8 & 61.8 & 62.6 & 33.1 & 33.3 & 61.3 & 58.5 & 36.9 & 58.0 & 48.9 & 53.2 & 50.4 & 44.0 & 59.0 & 59.3 & 49.9 \\
schizophrenia                             & 673  & 24.2 & 71.1 & 71.4 & 23.5 & 23.1 & 71.1 & 74.4 & 48.7 & 74.2 & 62.7 & 65.2 & 62.2 & 55.6 & 66.7 & 42.0 & 55.7 \\
psychotic disorder                        & 609  & 28.2 & 34.8 & 39.2 & 37.0 & 33.7 & 37.0 & 24.9 & 51.4 & 35.9 & 72.4 & 61.9 & 53.0 & 60.8 & 45.3 & 29.8 & 43.0 \\
schizoaffective disorder                  & 556  & 30.1 & 40.7 & 39.4 & 39.4 & 36.1 & 42.6 & 32.4 & 41.2 & 41.7 & 47.2 & 43.1 & 49.1 & 46.8 & 39.8 & 36.1 & 40.4 \\
unipolar depression                       & 480  & 30.4 & 59.8 & 63.1 & 31.8 & 31.5 & 61.9 & 67.3 & 50.0 & 68.5 & 62.2 & 70.5 & 69.0 & 60.1 & 61.9 & 32.7 & 54.7 \\
major depressive disorder                 & 475  & 36.1 & 55.1 & 55.5 & 38.8 & 35.7 & 55.1 & 67.8 & 55.9 & 64.8 & 62.1 & 69.6 & 75.3 & 70.5 & 59.5 & 55.1 & 57.1 \\
endogenous depression                     & 358  & 31.8 & 56.6 & 58.5 & 32.2 & 30.6 & 57.8 & 64.0 & 51.2 & 65.5 & 57.0 & 61.2 & 63.6 & 55.8 & 57.8 & 31.8 & 51.7 \\
obsessive-compulsive disorder             & 262  & 36.7 & 30.0 & 36.7 & 40.0 & 41.1 & 33.3 & 25.6 & 42.2 & 35.6 & 48.9 & 35.6 & 52.2 & 53.3 & 43.3 & 24.4 & 38.6 \\
anorexia nervosa                          & 187  & 26.7 & 32.0 & 32.0 & 40.0 & 36.0 & 34.7 & 21.3 & 40.0 & 24.0 & 46.7 & 48.0 & 54.7 & 52.0 & 41.3 & 37.3 & 37.8 \\
drug psychosis                            & 178  & 32.0 & 41.8 & 40.2 & 39.3 & 38.5 & 43.4 & 32.8 & 41.8 & 35.2 & 49.2 & 55.7 & 39.3 & 40.2 & 41.0 & 36.9 & 40.5 \\
manic bipolar affective disorder          & 163  & 41.7 & 43.7 & 43.7 & 47.6 & 42.7 & 49.5 & 49.5 & 55.3 & 49.5 & 63.1 & 62.1 & 65.0 & 66.0 & 50.5 & 47.6 & 51.8 \\
mixed anxiety and depressive disorder     & 135  & 56.9 & 21.6 & 21.6 & 62.7 & 56.9 & 23.5 & 39.2 & 68.6 & 35.3 & 62.7 & 58.8 & 70.6 & 70.6 & 37.3 & 31.4 & 47.8 \\
neurotic depression                       & 133  & 49.1 & 42.1 & 38.6 & 42.1 & 40.4 & 42.1 & 36.8 & 42.1 & 40.4 & 70.2 & 73.7 & 63.2 & 59.6 & 49.1 & 33.3 & 48.2 \\
\bottomrule
\end{tabular}
\begin{tablenotes}\scriptsize
\item Model abbreviations are identical to Table~\ref{tab:fine_grained_relations}.
\end{tablenotes}
\end{threeparttable}
}
\end{table*}

The fine-grained relation results in Table~\ref{tab:fine_grained_relations} reveal substantial variation across clinically salient relation families. The easiest relation on average is \texttt{indication}, with a mean accuracy of 59.1\%, followed by \texttt{disease\_disease} at 56.6\% and \texttt{exposure\_disease} at 55.7\%. In contrast, \texttt{contraindication} is by far the hardest relation, with a mean accuracy of only 35.8\%. This is notable because \texttt{contraindication} marks one of the most safety-sensitive boundaries in mental-health pharmacotherapy. Its difficulty helps explain why RP remains much weaker than ET or the small RT set, even for the strongest models.

The relation-level results also show that overall model quality does not imply uniform strength across relation families. GPT-4.1 is strongest on \texttt{contraindication} at 43.6\%, GPT-5-mini is strongest on \texttt{disease\_protein} and \texttt{exposure\_disease} at 65.9\% and 74.6\%, respectively, GPT-5.1-chat is strongest on \texttt{off-label use} at 70.9\%, and Qwen2.5-7B-Instruct is strongest on \texttt{indication} at 83.9\%. Qwen2.5-32B-Instruct achieves the best score on \texttt{disease\_disease} and \texttt{disease\_phenotype\_positive}. These row-wise inversions reinforce the idea that performance is relation-dependent rather than uniformly ordered by overall benchmark average.

The fine-grained entity results in Table~\ref{tab:fine_grained_entities} show that high benchmark incidence does not guarantee ease. Among the Top-15 high-incidence mental-health entities, the highest mean accuracies are observed for \emph{major depressive disorder} (57.1\%), \emph{schizophrenia} (55.7\%), and \emph{unipolar depression} (54.7\%). However, several prominent entities remain difficult, most notably \emph{anxiety disorder}, which has the highest benchmark incidence in this subset but only 40.1\% mean accuracy. \emph{Psychotic disorder} (43.0\%) and \emph{schizoaffective disorder} (40.4\%) are also challenging despite their high incidence. At the lower end, \emph{anorexia nervosa} (37.8\%) and \emph{obsessive-compulsive disorder} (38.6\%) are among the hardest entities in this set.

Taken together, the fine-grained analyses suggest that model failures are concentrated in clinically important and diagnostically heterogeneous parts of the graph. High-incidence depressive disorders are often handled reasonably well by the stronger models, but anxiety-spectrum, psychotic-spectrum, and medication-boundary distinctions remain much less stable.

\end{document}